\documentclass[a4paper, 10 pt, twocolumn]{IEEEtran}
\usepackage{cite}
\usepackage{amsmath,amssymb,amsfonts}
\usepackage{algorithm,algorithmic}
\usepackage{graphicx}
\usepackage{textcomp}
\def\BibTeX{{\rm B\kern-.05em{\sc i\kern-.025em b}\kern-.08em
    T\kern-.1667em\lower.7ex\hbox{E}\kern-.125emX}}
\usepackage{amsmath,amsfonts,amssymb}
\usepackage{epsfig,dsfont,amssymb,amsthm,amsbsy,mathrsfs,color} 
\usepackage{dcolumn}   
\usepackage{bm}        
\usepackage{subfigure} 
\usepackage{lineno,hyperref}  
\usepackage{cases}     
\usepackage{multirow}  
\usepackage{array}
\usepackage{comment}

\renewcommand{\figurename}{Fig.}

\begin{document}
\title{Convolutional Recurrent Neural Networks \\ for Glucose Prediction}
\author{Kezhi~Li, John Daniels, Chengyuan~Liu,  Pau~Herrero, Pantelis~Georgiou\\
Department of Electronic and Electrical Engineering, Imperial College London, London SW7 2AZ, UK\\

}

\maketitle

\begin{abstract}
Control of blood glucose is essential for diabetes management. Current digital therapeutic approaches for subjects with Type 1 diabetes mellitus (T1DM) such as the artificial pancreas and insulin bolus calculators
leverage machine learning techniques for predicting subcutaneous glucose for improved control. Deep learning has recently been applied in healthcare and medical research to achieve state-of-the-art results in a range of
tasks including disease diagnosis, and patient state prediction among others. In this work, we present a deep learning model that is capable of forecasting glucose levels with leading accuracy for simulated patient
cases (RMSE = 9.38$\pm$0.71 [mg/dL] over a 30-minute horizon, RMSE = 18.87$\pm$2.25 [mg/dL] over a 60-minute horizon) and real patient cases (RMSE = 21.07$\pm$2.35 [mg/dL] for 30-minute, RMSE = 33.27$\pm$4.79\% for
60-minute). In addition, the model provides competitive performance in providing effective prediction horizon ($PH_{eff}$) with minimal time lag both in a simulated patient dataset ($PH_{eff}$ = 29.0$\pm$0.7 for 30-min
and $PH_{eff}$ = 49.8$\pm$2.9 for 60-min) and in a real patient dataset ($PH_{eff}$ = 19.3$\pm$3.1 for 30-min and $PH_{eff}$ = 29.3$\pm$9.4 for 60-min). This approach is evaluated on a dataset of 10 simulated cases
generated from the UVa/Padova simulator and a clinical dataset of 10 real cases each containing glucose readings, insulin bolus, and meal (carbohydrate) data. Performance of the recurrent convolutional neural network is
benchmarked against four algorithms. The proposed algorithm is implemented on an Android mobile phone, with an execution time of $6$ms on a phone compared to an execution time of $780$ms on a laptop.
\end{abstract}


\linespread{0.97}

\section{Introduction}\label{Section:1}
Diabetes is a chronic illness characterised by the absence of glucose homeostasis. A healthy pancreas dynamically controls the release of insulin and glucagon hormones through the $\alpha$-cells and $\beta$-cells
respectively, in order to maintain euglycaemia \cite{NationalDia-ClassDia1979}. In Type 1 diabetes, an autoimmune disease, the $\beta$-cells are compromised and therefore suffer from impaired production of insulin. This
leads to periods of hyperglycaemia ( persistent blood glucose (BG) concentration $>180 mg/dL$) and hypoglycaemia ( {BG} concentration $< 70 mg/dL$)\cite{Facchinetti-AnOnline2013, Zavitsanou-InSilico2015}. Insulin therapy
is needed to maintain  {BG} levels in the advised target range \cite{Vettoretti-Type1Dia2018}.

 The standard approach to diabetes management requires people actively taking  {BG} measurements a handful of times throughout the day with a finger prick test - self monitoring of blood glucose. The recent development
 and uptake of continuous glucose monitoring (CGM) devices allow for improved sampling (5 minutes) of glucose measurements  { \cite{Ahmadi-AWireless2009}}. {This approach has proven to be effective in controlling BG and
 thus improving the outcome of subjects in clinical trials \cite{Facchinetti-ConGlu2016}. Further improvement of glucose control can be realised through prediction, which allows users to take actions ahead of time in
 order to minimise the occurrence of adverse glycaemic events. The challenges lie in multiple factors that influence glucose variability, such as insulin variability, ingested meals, stress and other physical activities
 \cite{Oviedo-AReview2017}.} In addition, {individual glycaemic responses are conditioned by high subject variability\cite{Vettoretti-Type1Dia2018,Pesl-AnAdvBolus2016}, leading to different responses between individuals
 under the same conditions.}

 {Machine learning (ML) allows intelligent systems to build appropriate models by learning and extracting patterns in data. The models discover mappings from the representation of input data to the output. Performances of
 traditional machine learning algorithms} such as logistic regression, k-nearest neighbours \cite{Karegowda-CascadingK2012}, or support vector regression \cite{Georga-MultiPred2013} heavily rely on the representation of
 the data they are given. { Typically, the features - information the representation comprises - are engineered with prior knowledge and statistical features (mean, variance) \cite{Yan-BloodGlucosePred2014}, principal
 component analysis (PCA) \cite{Abdi-PCA2010} or linear discriminant analysis \cite{Polat-ACascade2008}.}  Artificial neural networks (ANN) are also investigated widely in diabetes management
 \cite{perez-ArtiNN2010,Zecchin-NNIncor2012,Plis-AMachine2014,Mhaskar-ADeepLearningApp2017,OhioT1DM-dataset}.  {One advantage of ANN is that, normally no handcraft feature finding is required.} {However, ANN in the
 literature are mostly implemented fewer than 3 layers, hence its learning capacity is limited due to the model complexity.}
 {Deep learning, which incorporates multi-layer neural networks, has lead to significant progresses in computer vision \cite{Jia-Caffe2014}, diseases diagnosis \cite{Litjens-DLasTool2016}, and healthcare
 \cite{Miotto-DLforHealthcare2017,Zhu-DLAlg2018}.
Deep learning shows superior performance to traditional ML techniques due to this ability to automatically learn features with higher complexity and representations
\cite{Bengio-StatisticalLang2013,Schmidhuber-DLinNN2015,Li-RecurrentNeu2017,Zhang-InterpretableConv2018}. As a result, it encodes features that might not be previously known to researchers.}

In this paper, we propose a deep learning algorithm for glucose prediction using a multi-layer convolutional recurrent neural network (CRNN) architecture. The model is primarily trained on data comprising CGM,
carbohydrate and insulin data. After preprocessing, the time-aligned multi-dimensional time series data of BG, carbohydrate and insulin (other factors also can be considered) are fed to  { CRNN} for training. The
architecture of the CRNN is composed of three parts: a multi-layer convolutional neural network that extracts the data features using convolution and pooling, followed by a recurrent neural network (RNN) layer with long
short term memory (LSTM) cells and fully-connected layers. {The model is trained end-to-end.
The convolutional layer comprises a 1D Gaussian kernel filter to perform the temporal convolution, and pooling layers are used for reducing the feature set.} A variant of LSTM model is leveraged since LSTM shows good
performances in predicting time series with long time dependencies \cite{Goodfellow-DL2016}.
The final output is a regression output by fully connected layers.  {The CRNN model} is realized using the open-source software library Tensorflow \cite{tensorflow2015-whitepaper}, and it can be easily implemented to
portable devices with its simplified version Tensorflow Lite. The performance of the proposed method is evaluated on datasets of simulated cases as well as clinical cases of T1DM subjects, and compared against {benchmark}
algorithms including support vector regression (SVR) \cite{Georga-MultiPred2013}, the latent variable model (LVX) \cite{Chunhui-PreSub2012}, the autoregressive model (ARX) \cite{finan2009experimental}, and neural network
for predicting glucose algorithm (NNPG)\cite{perez-ArtiNN2010}. The results show the competitive performance of the method.

 {As far as we know, the proposed method is a pioneering work in glucose forecast implementing deep neural networks that incorporates the merits of both CNN and RNN, and we modify them to suit the task of glucose
 prediction. It achieves competitive performances in terms of forecasting accuracy comparing to benchmark methods, with superior performances in terms of RMSE. It applies multi-layer NN in smart phones with applications
 in diabetes management. The paper is organized as follows. Section II briefly introduces the properties of the glucose data. Section III addresses the method and architecture of the proposed convolutional recurrent
 neural network. The performance of the proposed method is evaluated and discussed in Section IV. Section V discusses important details of the method. Finally Section VI summaries the paper.}

\section{ {Data Acquisition and Setting}}
\subsection{ {Data Acquisition}}
 {The data used in this paper include two datasets, \textit{in-silico} data and clinical data. \textit{In-silico} data consisting  of 10 adult T1DM subjects was generated using the UVA/Padova T1D, which is a simulator for
 glucose level simulation approved by the Food and Drug Administration (FDA) \cite{Man-UVA/PADOVA2014}.  In this work, we used a modified version of the simulator which includes such variability. In particular, the
 variability on meal composition, insulin absorption, carbohydrate estimation and absorption, and insulin variability were included. In addition, a simple model of physical exercise was also used. Details about how the
 simulator was modified can be found in \cite{herrero2017enhancing}.

Clinical data was obtained from a clinical study at Imperial College Healthcare NHS Trust St. Mary’s Hospital, London (UK) consisting of multiple phases evaluating the benefits of an advanced insulin bolus calculator for
T1D subjects \cite{Reddy-CliSafety2016}. The dataset in consideration was collected from a 6-month period involving 10 adult subjects with T1D.
The information included in the dataset comprises glucose, meal, insulin, and associated time stamps. In building the dataset, we mainly consider CGM and self-reported data such as insulin boluses, meals, and exercise
similar to the in-silico dataset. Before that, we exclude
participants whose data exhibited large gaps (corresponding to weeks of missing data), insufficient reports of exercise over the 6-month period, and extensive errors in sensor readings.

The CRNN model can be applied to datasets where other inputs are available, such as self-reports of exercise, stress and alcohol consumption. We believe that these information are useful and will increase the forecast
accuracy in some cases. But in this section we only consider CGM data recorded every 5 minutes, meal data indicating meal time and amount of carbohydrates, as well as insulin data with each bolus quantity and the
associated time as input in the model.
}

\subsection{ {Data Setting}}
 {
The \emph{in silico} data considered in this section is generated via UVA/Padova T1D. This simulator serves as robust and validated framework for generating simulated cases. The cohort of T1D cases generated can be
configured with varying meal and insulin information such that each case sufficiently differs.
We generate a dataset of 10 unique adult cases and each has 360 days of data for each case. There are 3 meals per day. Insulin entries vary in each day, from 1 to 5. The insulin entry can be with a meal (meal and insulin
at almost the same time), or without a meal (correction bolus).
A simple exercise model is considered at certain points, which occur occasionally at any time except for nights. The training data accounts for 50\% of the dataset, and the testing set is the rest of data.

The clinical data was collected from T1DM subjects in a 6 month clinical trial.
The CGM data was measured using Dexcom G4 Platinum CGM sensors , with measurements received every 5 minutes. The CGM sensors were inserted from the first day of the study, and calibrated according to the manufacturer
instructions. Other information available in the dataset such as meal, insulin, exercises was logged by the diabetic subjects themselves.
Though the selected data has good quality, many periods of missing data, bad points or unexpected fluctuations exist. Similar to the \emph{in silico} experiment, each subject's clinical data is halved for training and
testing data.
}

 \begin{figure*}[t]
  \includegraphics[width=15.8cm,height=6.5cm]{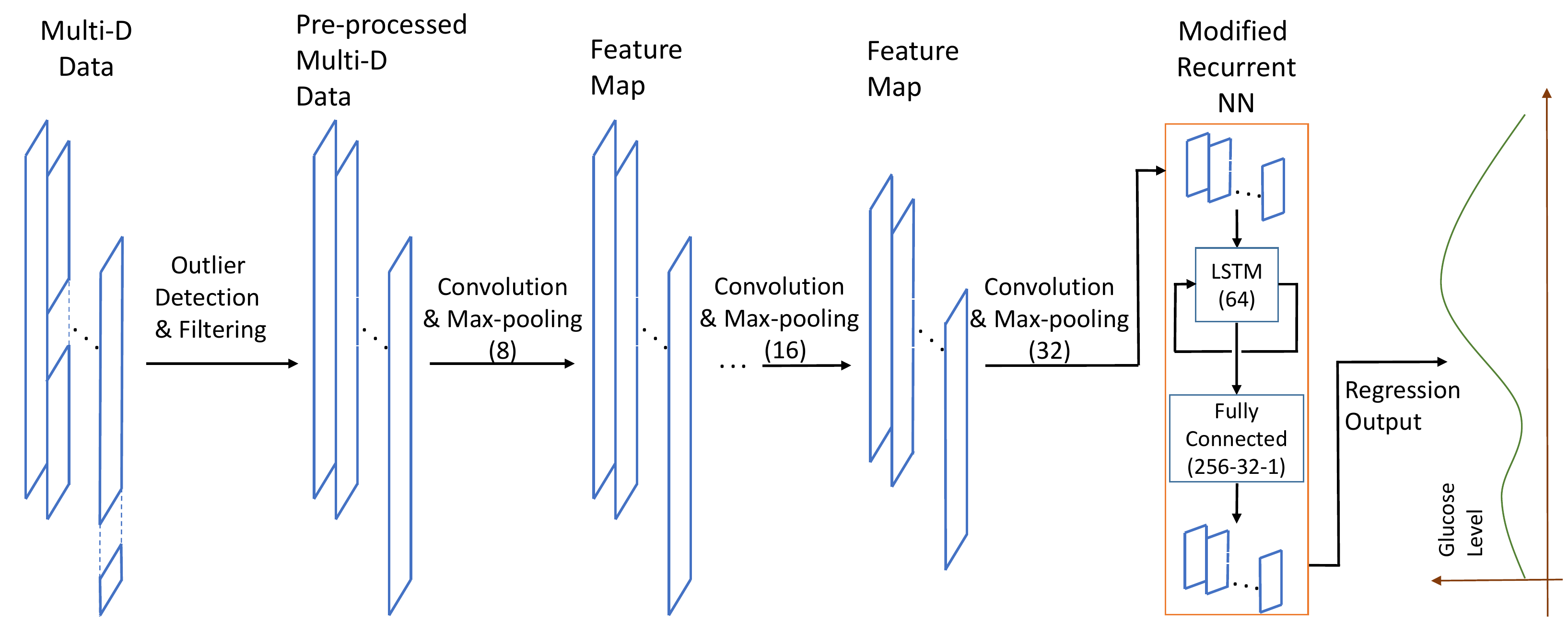} 
  \centering
  \caption{The architecture of the proposed convolutional recurrent neural network for  {BG} prediction. The data at the left is the concatenated time series data including glucose level, carbohydrate, insulin and other
  factors. After outlier filtering, the multi-dimensional data can be sent to the multi-layer convolutional component. Then the resultant time series is sent to the modified recurrent neural network component presented in
  a red frame, which includes LSTM cells and dense fully connected layer. Finally, the resultant is converted back from ``change of the glucose value'' to    ``absolute glucose value''. The output is the future glucose
  values of PH (eg. PH = 30 mins).}\label{fig:cnn_lstm1}
  \centering
 \end{figure*}

\section{Methods}
After introducing the data acquisition and setting, the method is explained explicitly.
The approach consists of several components: preprocessing, feature extraction using CNN, time series prediction using LSTM and a signal converter. The architecture of the proposed CRNN is shown in Fig.
\ref{fig:cnn_lstm1}. In the diagram, the input of the algorithm is time series of glycaemic data from CGM, carbohydrate and insulin information (time and amount); other related information are optional (exercises,
alcohol, stress, etc.). The output of preprocessing is cleaned as time-aligned glycaemic, carbohydrate and insulin data, which are then fed to the CNN.
The output of CNN serves as the input of RNN, which is a multi-dimensional time series data, representing the concatenation of features of the original signals. The output of the RNN is the predictive BG level $30$-min
(or $60$-min) later, while hidden states are inherited and updated continuously internally inside of the RNN component. The model is trained end-to-end. We evaluate the models with $30$ and $60$-min prediction horizon
(PH) because it is widely used in glucose prediction software, and is easier to compare results with other works \cite{Sparacino-GluCon2007, Mougiakakou-NNbasedGlucose,Plis-AMachine2014, Georga-MultiPred2013,
Chunhui-PreSub2012,perez-ArtiNN2010}. We proceed with an explanation data pipeline and components of the model architecture.

\subsection{Outlier Detection and Filtering}

The main purpose of the preprocessing component is to clean the data, filter the unusual points and make it suitable as the input to the neural network. Besides the normal steps including time stamp alignment and
normalization,
the most important operation to improve the data quality is the outlier detection, interpolation/extrapolation and filtering, in particular for clinical data. Because in clinical data, there are many missing or outlier
data points due to errors in calibration, measurements, and/or mistakes in data collection and transmission. Here, several methods can be used to handle these scenarios \cite{Atanassov-AlgorithmsForOpt2009}. They include
 dimension reduction model to project data into lower dimensions \cite{Shum-PCAwihtMissing2001}, proximity-based model to determine the data by cluster or density \cite{Li-TowardsMissing2004}, and probabilistic stochastic
 filters \cite{Lekha-RealTimeNon2018}  {to rule out} outliers.

For some cases when the data fluctuates with high frequency, 1D Gaussian kernel filter {can be implemented on the glucose time series. A smoothed continuous time series of glycaemic data can be obtained, along with} the
time-aligned carbohydrate and insulin information.
 In this paper, for \emph{in silico} data we do not use filters  because the dataset is already clean. For clinical data used in this paper, we use the Gaussian filter. The 3-dimensional time series that covers the last
 $2$ hours before the current time is sent to the neural network as input. A sliding window of size $24$ is determined to train the model.  {Because during the experiments we find that $24$ is an optimal setting
 considering the tradeoff between the prediction accuracy and the computation complexity. In \cite{perez-ArtiNN2010} the NNPG algorithm uses a similar window size of $20$. }

\subsection{A Multi-layer Convolutional Network}

\begin{figure}[h]
  \includegraphics[width=8.9cm]{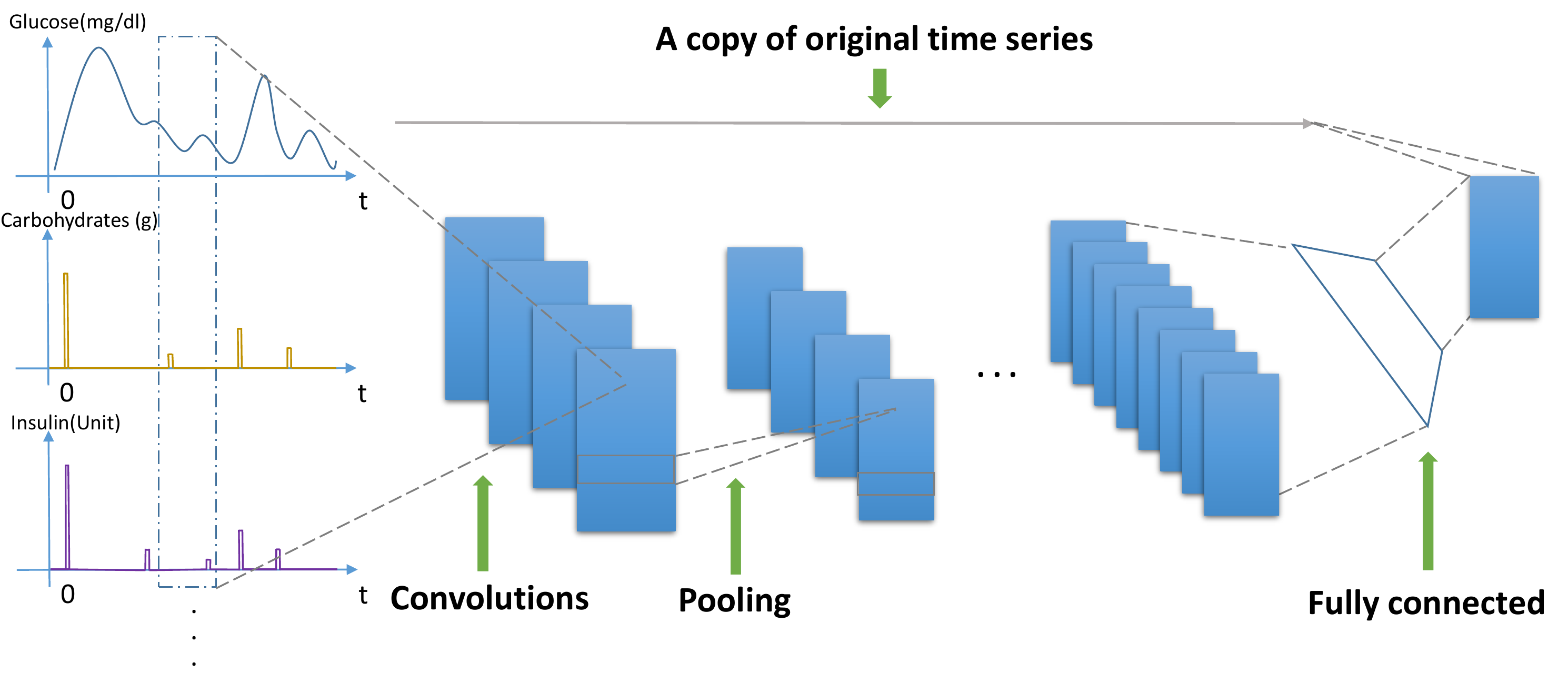} 
  \centering
  \caption{An illustration of the proposed CNN architecture. The multi-dimensional aligned time series data are concatenated, and then sent to a multi-layer CNN composed of convolutional layers and pooling layers.
  Finally, after going through a fully connected layer, the final output is the summation of the model output and a copy of the original CGM time series.  }\label{fig:cnn_1}
  \centering
 \end{figure}

The filtered time series signal goes through the multi-layer convolutions,  {which transform} the input data into a set of feature vectors. The convolution operation follows the temporal convolution definition in which
\begin{equation}\label{eq:conv}
  z[m] = \sum_{i = -l}^{l} x[i] \cdot \delta[m-i],
\end{equation}
where $x$ represents the input signal, $\delta$ denotes the kernel,  {$z$ is the result of the convolution, and $m$ is the result's index. Specifically in the first layer, $x$' length is the sliding window size of $24$,
kernel $\delta$ has a size of $8$.  The dimension of data in each layer can be referred to the Appendix.}
The input signal can be fed
using a sliding window setting. The windows can be overlapped or non-overlapped, determined by the allowed CNN size and computations. The CNN automatically learns the associated weights and recognises particular patterns
and features in the input signal that can best represent the data for future time steps.
It is illustrated in Fig. \ref{fig:cnn_1}.

  \begin{figure*}[t]
  \includegraphics[width=15cm]{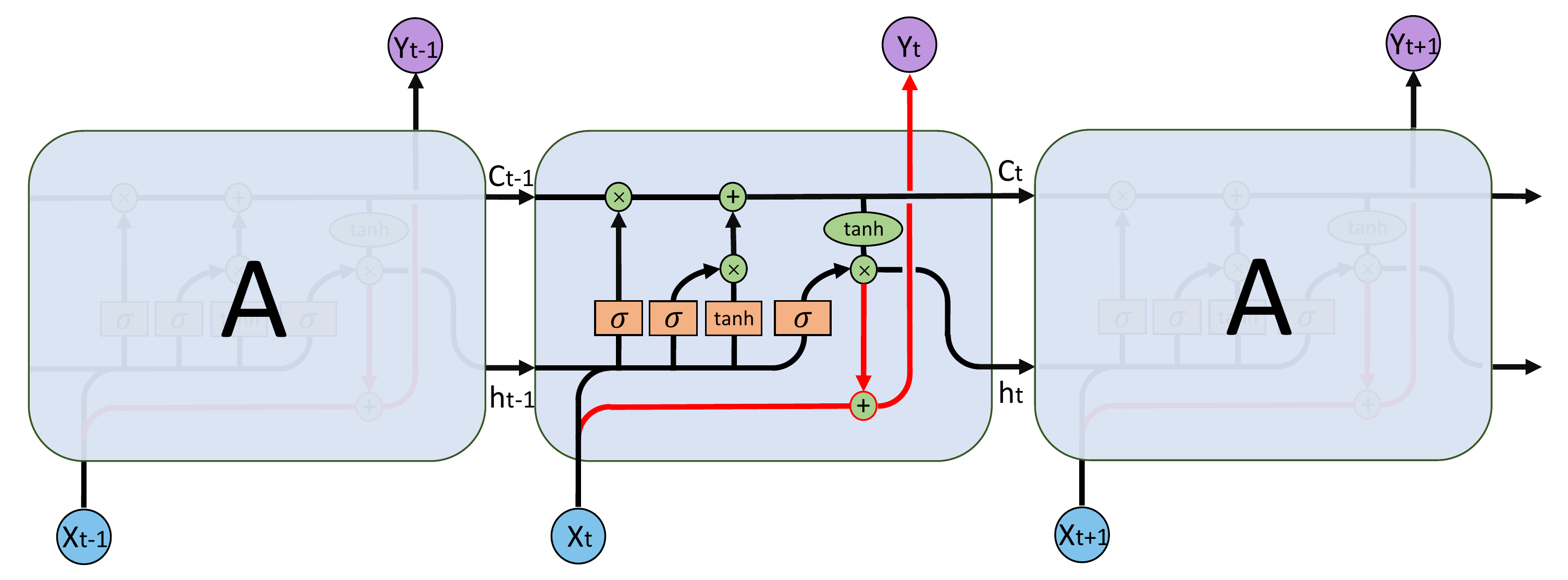} 
  \centering
  \caption{An illustration of the proposed modified recurrent neural network in the inference. It origins from the conventional LSTM cell, including an input gate, an output gate and a forget gate. The difference lies in
  the output part, which indicates in red colour in the figure. The output of $h_t$ is still the internal parameter which is transmitted to the next cell. However, the output value signal is $y_t$ instead of $h_t$. It is
  because in the training process, the target of RNN is the change of the glucose value between current time and 30 mins later. Thus the predictive glucose level after 30 mins should be the inference value plus the
  current glucose value.}\label{fig:lstm_chain}
  \centering
 \end{figure*}

The proposed CNN  consists of 3 convolutional layers, with max pooling applied to down-sample the feature map obtained from the previous convolutional layer. It is common to periodically insert a pooling layer in-between
successive convolutional layers to progressively {reduce the size of the representation, as well as the computation}. It also guards against the problem of overfitting. For instance, if the accepted size is $L1 \times
D1$, and the down-sampled parameters are spatial extent $F$ and stride $S$, and it results in a max-pooling vector $Y$ of size $L2 \times D2$ where
\begin{equation}\label{eq:LD}
L2 = (L1-F)/S + 1; \ D2 = D1;  \ Y_i = \max(y_i^*)
\end{equation}
where $y_i^*$ is the vectors after being down-sampled, $Y_i$ is the feature map and $\max()$ is the operator that computes the maximum value.  During the training process, the CNN is trained by back-propagation and the
stochastic gradient descent method. The initial weights of the network are set randomly, and the mean-absolute-error is set as the cost function to be minimised in the training. Partial derivatives of the error in terms
of the weights $w_i$ and bias $b_i$ are computed, and the associated $w_i$ and $b_i$ are updated accordingly. The last convolutional layer feeds directly into the recurrent layer that makes up the next component in the
architecture.

\subsection{A Modified Recurrent Layer}

 An LSTM network comprised of 64 LSTM cells is adopted as recurrent layers \cite{Hochreiter-LongShort}. Each LSTM cell consists of an input gate, an output gate and a forget gate. Each of the three gates can be thought of
 as a neuron, and each gate achieves a particular function in the cell. The LSTM network is good at building predictive models for time series and sequential data \cite{Graves-SpeechRecognition}. These cells retain
 previous data patterns over arbitrary time intervals, thus the internal ``memory'' can predict the future output according to the previous states. Its memory  {can be} updated simultaneously when new data are fed to the
 model.

The output of the CNN, {a multi-dimensional time series, is connected to the LSTM network. We generated an RNN with $1$ hidden layer}, consisting of a wide LSTM layer consisting of $64$ cells. A dropout applied after the
LSTM layer. Dropout refers to ignoring neurons randomly during the training phase. It has been verified that in many cases that dropout can effectively avoid overfitting problems and improve the generalisation
\cite{Srivastava-DropoutASimple}.

The main difference between a normal LSTM and the proposed LSTM is that the proposed model {has a transform and a recovery step. They modify BG values before and after the conventional LSTM. In training phase}, instead of
BG values directly, we use the changes of BG between the current BG $x(t)$ and the future BG $x(t+6)$ as target labels. It is called the transform step. The input sliding window matrices are multi-dimensional time series
(including BG values, meal, insulin). After the model has been trained, the inference output is the change of BG $\triangle x(t)$ between $x(t)$ and $x(t+6)$. Thus the prediction of BG at time $t+6$ can be calculated as
$x(t+6) = x(t) + \triangle x(t)$. This is called the recovery step. 

Specifically,  a modified LSTM estimates the conditional probability $p(y_1, \cdots,y_{T'}|x_1, \cdots,x_{T})$ given a sequence of data, where $x_1, \cdots,x_{T}$ denotes the input sequence and $y_1, \cdots,y_{T'}$ is the
corresponding output sequence, with size $T$ and $T'$ respectively. If $x_t, h_t, c_t$ are used to denote the input vector, output vector and memory cell vector respectively; $W, U,$ and $b$ are the parameter
matrices/vectors that can be learned in the network. $f_t, i_t,$ and $o_t$ denote the forget gate, input gate, and output gate vectors. Then mathematical form of the update process can be explicitly written as
\begin{equation}\label{eq:lstm_update}
\begin{split}
f_t &= \sigma_g(W_fx_t + U_fh_{t-1}+b_f) \ \ i_t = \sigma_g(W_ix_t + U_ih_{t-1}+ b_i)  \\
o_t &= \sigma_g(W_ox_t + U_oh_{t-1}+b_o)  \ \  g_t = \sigma_t(W_gx_t + U_gh_{t-1}+ b_g)  \\
c_t &= f_t \circ c_{t-1} + i_t \circ \sigma_t(g_t)  \ \ \ \ \ \ \ \ \ \ \ \  h_t = o_t \circ \sigma_t(c_t), \\
y_t &= o_t \circ \sigma_t(c_t) + x_t,
\end{split}
\end{equation}
where $\sigma_g, \sigma_t, \circ$ is the sigmoid function, hyperbolic tangent and entrywise product, respectively. In (\ref{eq:lstm_update}), the 1st to 5th equations are the same to the equations of normal LSTM. However
the last 2 equations of $h_t$ and $y_t$ are modified accordingly.
 It has been shown in Fig. \ref{fig:lstm_chain}, where the difference between the proposed modified LSTM and the normal LSTM is indicated in red. The output of the LSTM cell is not $h_t$, but $y_t$ in inference. The $h_t$
 is used as an internal hidden state that goes to the next time step. The output $y_t$ is calculated from $h_t$ plus the original input $x_t$. That is because the output (and targets) of the neural network is the change
 of the  {BG} level. The value of the predictive glucose level needs to be recovered from the glucose change by adding the baseline glucose value.

Finally, the last layer of RNN feeds a multi-layer fully connected network, which consists of 2 hidden layers (256 neurons and 32 neurons) and an output layer (a single neuron) with the glucose change as output. The fully
connected layer produces the output with an activation function
\begin{equation}\label{eq:fully}
 Z_i = act(\sum_{i=1}^{N} Y_i w_i + b_i),
\end{equation}
where $Z_i$ is the multi-dimensional output, $act()$ is an activation function, $w_i$ and $b_i$ are weights of the fully connected network. Particularly, $act()$ can be chosen from a set of activation functions such as
sigmoid function $act(a)= 1/(1+e^{-a})$, rectifier $act(a) = \log{(1+\exp(a))}$ or simple linearly $act(a) = a$. In this paper we choose {the linear function  $act(a) = a$ as the activation function for its simplicity}.
 In the training, a gradient descent optimisation is used.
 The mean absolute error between the target and the predictive value is being minimised. The optimiser we use is \emph{RMSprop}, because it is a good choice for recurrent neural networks.  { It usually maintains a moving
 (discounted) average of the square of gradients \cite{Ruder-AnOverviewOf2016}, and divides gradient by the root of this average.}

\subsection{Software and Hardware}
\begin{figure}[t]
  \includegraphics[width=8cm]{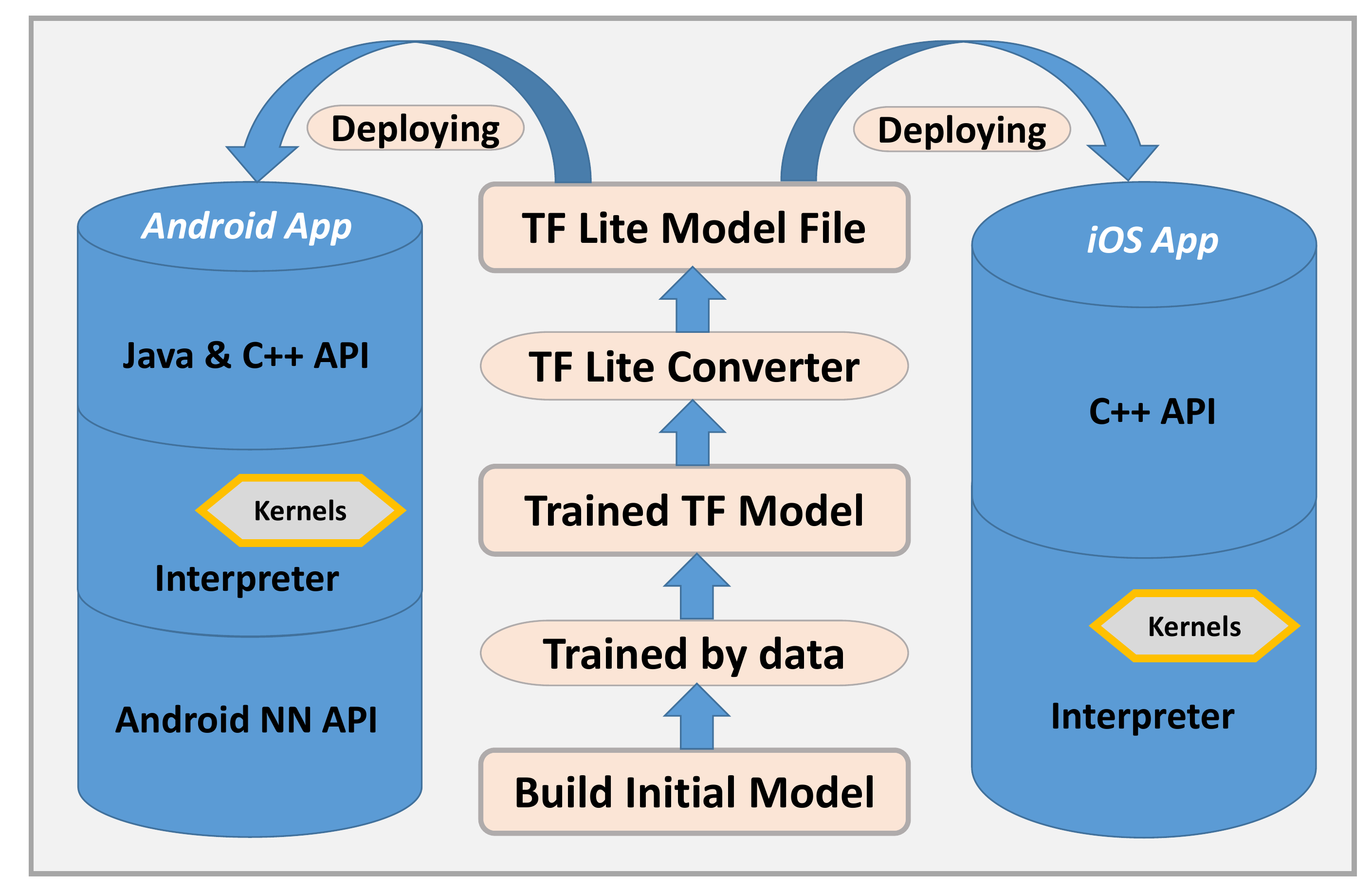} 
  \centering
  \caption{ A flow diagram explains the procedure to implement deep learning models as Tensorflow Lite files to portable devices. In the figure, yellow square frames denote the models or files obtained after each
  operation, and yellow elliptic frames denote the associated operations applied. After TF Lite model files are created, they can be deployed in Android or iOS app with slightly different settings as shown in the
  cylinders. }\label{fig:flow1}
  \centering
 \end{figure}

After the model has successfully undergone training and validation, we implement our algorithm on mobile phones through Tensorflow Lite due to its efficiency running on portable devices. The model is converted to a Lite
model file and installed on an Android or iOS system. It needs the associated application programming interface (API) and interpreter to carry out the inference. Fig. \ref{fig:flow1} shows how the Tensorflow Lite model
file \cite{tensorflow2015-whitepaper} is wrapped and loaded in a mobile-friendly format.

CGM sensors were used to collect data. We utilize computers equipped with graphics processing unit (GPU) to carry out the training and inference processes. In practice an Intel i7-7700K CPU with 4.20 GHz, 32.0 GB memory
with GPU NVIDIA\@ GeForce GTX 1080 Ti were used in the experiments. The program was written in Python 3.6, using CUDA 9.0 to perform the parallel computing.
Tensorflow architecture was implemented, because its compatibility and merits for large-scale distributed training and inference. 

\section{Results}

 In this section we test the proposed CRNN algorithm for glucose level forecasting using a \emph{in silico} dataset and a clinical dataset. The performance of the proposed algorithm is contrasted with that of four
 baseline methods: NNPG, SVR, LVX and ARX (3rd order). The results are compared with the same input data after the same pre-processing. The performance of the methods are compared based on the accuracy over 30- and
 60-minute prediction horizon. In addition, we evaluate the time lag of the prediction.
 Different algorithms were tested on the \emph{in silico} data generated in a way described previously.
The parameters involved in these algorithms are tuned carefully for optimal result. In SVR, the SVR function in Python is applied with the optimal parameters ($C=1e2, \gamma=0.01,cache_{size}=1000$). The LVX method is
applied based on the MATLAB code provided in \cite{Chunhui-PreSub2012}, the optimal predictor length and the number of LVs are $J_x = 4$ and $N_{LV} = 4$ respectively. This represents 20-minute historical data of glucose
measurement, insulin and meal information being used for prediction. The 3rd order ARX model is optimized by MATLAB function $arx()$ for every specific subject.

\subsection{Criteria for Assessment}
 {Several criteria are used to test the performance of the proposed algorithm. The root-mean-square error (RMSE) and mean absolute relative difference (MARD) between the predicted and reference glucose readings serve as
 the primary indicators} to evaluate the accuracy.

\begin{equation}\label{eq:RMSE}
  RMSE = \sqrt{\frac{1}{N}\displaystyle\sum_{k=1}^{N}(y(k)-\hat{y}(k|k-PH))^2},
\end{equation}
where $\hat{y}(k|k-PH)$ denotes the prediction results provided the historical data and $y$ denotes the reference glucose measurement, and $N$ refers to the data size.
\begin{equation}\label{eq:MARD}
  MARD = \frac{1}{N}\sum_{k=1}^{N}\frac{|\hat{y}_k(k|k-PH)-y(k)|}{y(k)}.
\end{equation}
The RMSE and MARD provide an overall indication of the predictive performance.
As mentioned earlier, the benefit of glucose prediction is avoiding adverse glycaemic events. In the clinical context, these metrics are limited in the insight they provide. Additional metrics are needed to assess the
proposed algorithm in the following perspective:
\begin{description}
	\item[$\bullet$] Capability of the forecasting algorithm in differentiating between adverse glycaemic events and non-adverse glycaemic events.
	\item[$\bullet$] Time delay in the predicted glucose readings and reference values to evaluate the response time provided to deal with the potential adverse glycaemic event.
\end{description}

The Matthews Correlation Coefficient (MCC)
is used to evaluate the performance of the algorithms for detecting either adverse glycaemic event (hypoglycaemia or hyperglycaemia).

\begin{equation}\label{eq:AccSen}
   MCC \!=\! \frac{(TP\times TN)\!-\!(FP\times FN)}{\sqrt{(TP\!+\!FP)(TP\!+\!FN)(TN\!+\!FP)(TN\!+\!FN)}} ,
  \end{equation}
where $TP, FP, FN, TN$ stand for true positive, false positive, false negative, and true negative respectively. Here, positive indicates a hypoglycaemia ($<70$ mg/dL)/hyperglycaemia ($>180$ mg/dL) event in the next or
previous 30 or 60 minutes, and true means that the classification is correct.  We consider a true adverse event to have occurred when either scenario persists in the CGM data for at least 20 minutes
\cite{international_hypoglycaemia_study_group_glucose_2017}. In addition, we consider an event a True Positive when the predicted event is at most 10 minutes (PH+2 timesteps ahead) leading or within 25 minutes of the
prediction horizon lapsing (1 timestep for PH = 30 min, and 7 timesteps for PH = 60 min) the reference event.
A standard confusion matrix typically includes the Accuracy as opposed to Matthews Correlation Coefficient (MCC). This modification addresses the imbalance in classes inherent in this situation - non-adverse events far
outweigh adverse events.

\begin{figure*}[!ht]
     \centering
             \includegraphics[width=4.9in,height=3.2in]{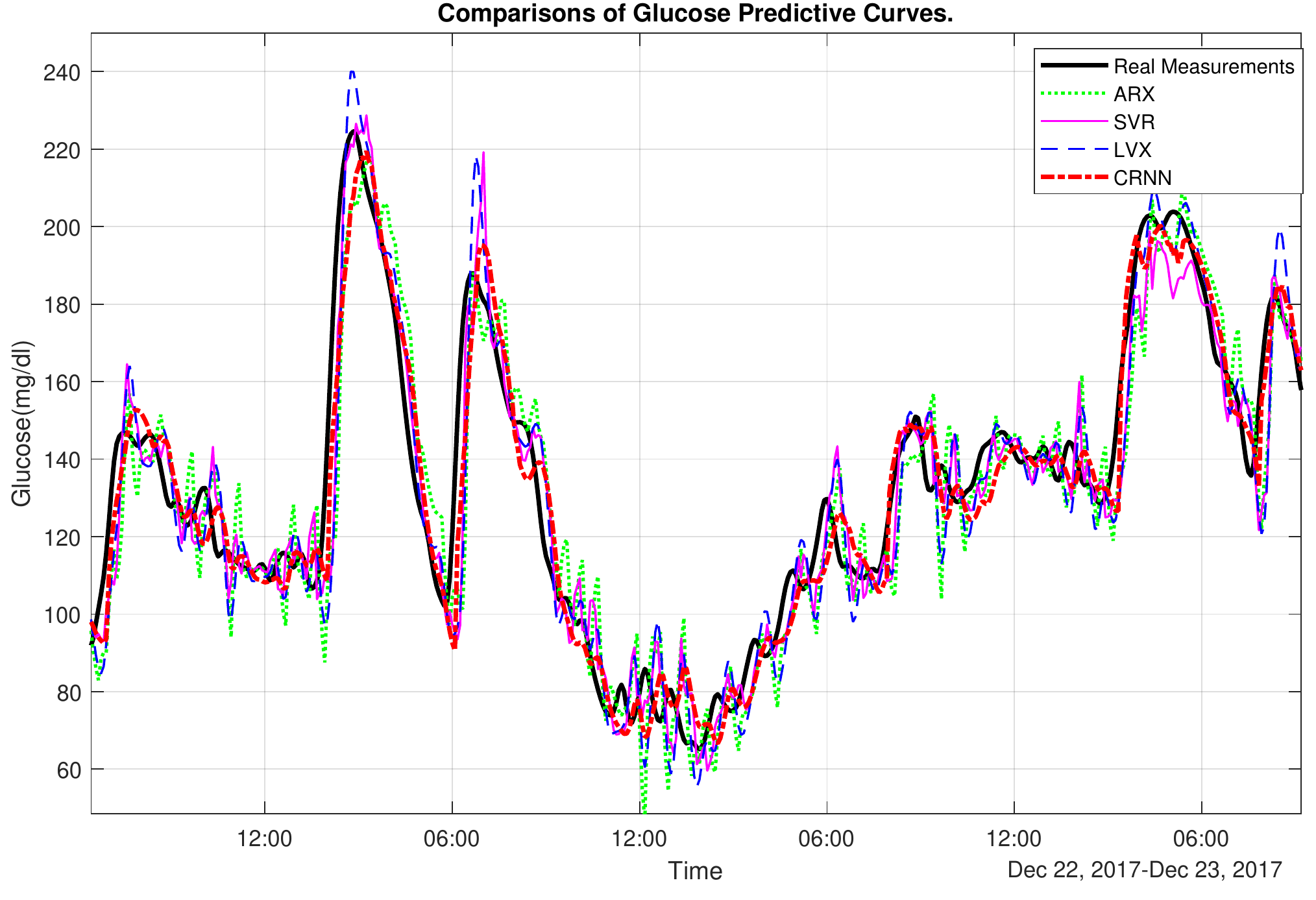}
             \caption{One-day period prediction results for virtual adult $4$. The solid black line, dotted green line, solid magenta line, dashed blue line, dash-dotted red line indicate the simulated glucose
             measurements, the prediction results of the 3rd order ARX method, the prediction results of the SVR method, the prediction results of the LVX algorithm, the prediction results of the CRNN method,
             respectively. }
             \label{fig:C1}
\end{figure*}

The effective prediction horizon is defined as the prediction horizon, taking into account delays due to the responsiveness of the algorithm for a predicted value relative to its reference value. Cross correlation of the
predicted and actual readings is employed in performing a time delay analysis of the proposed algorithm to determine the effective prediction horizon.

\begin{equation}\label{eq:Delay}
\begin{split}
PH_{eff} &= PH - \tau_{delay} \\
         &= PH - \underset{\tau}{\arg\max} (\hat{y}_k(k|k-PH) \star y(k)).
\end{split}
\end{equation}

A singular quantitative metric is not sufficient in evaluating performance of the proposed algorithm. Consequently, the set of metrics collectively give a comprehensive description of the quality of the prediction
algorithm performance.
 {$p$-values are calculated for other algorithms comparing to the proposed algorithm in terms of smaller RMSE, MARD or longer $PH_{eff}$. A Shapiro-Wilk Test is used to ascertain the normality of the results before
 performing a paired t-test to derive the $p$-values. Across all results, the tests show the null hypothesis (samples drawn from a Gaussian distribution) cannot be rejected.}

\begin{table}[!ht]
 {
	\small
	\caption{RMSE and MARD comparison of different prediction methods for 10 virtual adult diabetic subjects }\label{table:Adult10RMSEMARD}
	\begin{tabular}{|c|c|c|c|c|c|c|}
		\hline
		 \multirow{2}{*}{PH}  & \multirow{2}{*}{Metric} & CRNN & NNPG & SVR & LVX & ARX \\
		\cline{3-7}
		 \multirow{13}{*}{30} &&\multicolumn{5}{c|}{Overall}\\
		\hline
		        & RMSE  & $\textbf{9.38}$ & $12.91$\textsuperscript{$\ddagger$} &$12.48$\textsuperscript{$\ddagger$} & $11.32$\textsuperscript{$\dagger$} & $13.27$\textsuperscript{$\ddagger$}\\
		                       &(mg/dL)&$\pm$0.71&$\pm$1.19&$\pm$1.94&$\pm$1.34&$\pm$1.19\\
		\cline{2-7}
		                       & MARD  & $\textbf{5.50}$ & $7.05$\textsuperscript{$\ddagger$} &$6.40$  & $6.59$\textsuperscript{$\ddagger$} & $7.46$\textsuperscript{$\ddagger$}  \\
		                       & (\%)  &$\pm$0.62&$\pm$0.94&$\pm$1.36&$\pm$0.80&$\pm$1.02\\
		\cline{2-7}
		                       &       \multicolumn{6}{c|}{Hyperglycaemia}\\
		\cline{2-7}
		                       & MCC & $\textbf{0.84}$ & $0.83$ &$0.84$ & $0.83$ & $0.81$  \\
		                       &     & $\pm$0.05&$\pm$0.04&$\pm$0.04&$\pm$0.04&$\pm$0.05\\
		\cline{2-7}
		                       &     \multicolumn{6}{c|}{Hypoglycaemia}\\
		\cline{2-7}
		                       & MCC & $0.79$ & $0.64$ &$0.79$ & $\textbf{0.83}$ & $0.78$  \\
		                       &     & $\pm$0.15&$\pm$0.20&$\pm$0.10&$\pm$0.06&$\pm$0.10\\
		
		\cline{2-7}
		                       &      \multicolumn{6}{c|}{Effective Prediction Horizon}\\
		\cline{2-7}
		                       & Time & $\textbf{29.0}$ & $20.8$\textsuperscript{$\ddagger$}&$23.3 $\textsuperscript{$\ddagger$} & $27.5$\textsuperscript{$\dagger$} & $20.5$\textsuperscript{$\ddagger$}\\
		                       &(min) & $\pm$0.7&$\pm$1.8&$\pm$1.6&$\pm$1.3&$\pm$1.7\\
		\hline
		\multirow{13}{*}{60}&\multicolumn{6}{c|}{Overall}\\
		\cline{2-7}
		& RMSE  & $\textbf{18.87}$ & $24.24$\textsuperscript{$\ddagger$} &$23.46$\textsuperscript{$\ddagger$} & $22.42$\textsuperscript{$\dagger$} & $25.73$\textsuperscript{$\ddagger$}\\
		                       &(mg/dL)&$\pm$2.25&$\pm$3.01&$\pm$3.33&$\pm$2.74&$\pm$3.24\\
		\cline{2-7}
		                       & MARD  & $\textbf{9.16}$ & $13.70$\textsuperscript{$\ddagger$} &$10.83$  & $12.20$\textsuperscript{$\ddagger$} & $13.75$\textsuperscript{$\ddagger$}  \\
		                       & (\%)  &$\pm$1.16&$\pm$1.88&$\pm$1.48&$\pm$1.82&$\pm$2.45\\
		\cline{2-7}
		                       &       \multicolumn{6}{c|}{Hyperglycaemia}\\
		\cline{2-7}
		                       & MCC & $0.82$ & $0.79$ &$0.78$ & $\textbf{0.86}$ & $0.64$  \\
		                       &     & $\pm$0.05&$\pm$0.06&$\pm$0.07&$\pm$0.04&$\pm$0.05\\
		\cline{2-7}
		                       &     \multicolumn{6}{c|}{Hypoglycaemia}\\
		\cline{2-7}
		                       & MCC & $\textbf{0.80}$ & $0.38$ &$0.79$ & $0.80$ & $0.72$  \\
		                       &     & $\pm$0.14&$\pm$0.39&$\pm$0.10&$\pm$0.07&$\pm$0.12\\
		
		\cline{2-7}
		                       &      \multicolumn{6}{c|}{Effective Prediction Horizon}\\
		\cline{2-7}
		                       & Time & $\textbf{49.8}$ & $31.0$\textsuperscript{$\ddagger$}&$32.6 $\textsuperscript{$\ddagger$} & $44.2$\textsuperscript{$\ddagger$} & $19.8$\textsuperscript{$\ddagger$}\\
		                       &(min) & $\pm$2.9&$\pm$4.7&$\pm$4.1&$\pm$2.7&$\pm$2.7\\
		\hline
		\multicolumn{6}{r}{\textsuperscript{*}\scriptsize{$p$-value  $\leq$ 0.05}}  {\textsuperscript{$\dagger$}\scriptsize{$p$-value  $\leq$ 0.01}}  {\textsuperscript{$\ddagger$}\scriptsize{$p$-value $\leq$ 0.005}}\\
	\end{tabular}

\centering
}
\end{table}

\subsection{In Silico Data}
 {The results of RMSE, MARD and forecasting of adverse glyvaemic events are summarized in the Table \ref{table:Adult10RMSEMARD}.}
In the Table we compare the predictive error of the algorithms to measure the accuracy of the algorithms. {The CRNN algorithm exhibits the best overall RMSE and MARD for the 10 simulated cases at short(30) and long
term(60) predictions.} The results in Table \ref{table:Adult10RMSEMARD} are statistically significant relative to each algorithm. This observation is also evident in both the hyperglycaemia and hypoglycaemia region.  {In
the hyperglycaemia region the CRNN shows a statistically significant improvement in the glucose prediction
over most other algorithms, with the exception of LVX.} {CRNN reports a statistically significant improvement in effective prediction time (+1.5min for 30-min and +5.6min for 60-min) over LVX, and it is better than the
rest thus giving the user more time to take action.
On the whole CRNN can be evaluated as the best algorithm.}
The CRNN model also reports relatively low standard deviations from which we infer a benefit in building individualized models.

An illustration of a comparison of various algorithms for 30-minute shown in Fig. \ref{fig:C1} for a virtual adult 4.
As seen in Fig. \ref{fig:C1}, CRNN exhibits the best responsiveness as the predictive curve responds rapidly towards the sharp glycaemic uptrend. The algorithm learns representations that appropriately account for both
sharp slopes and gradual increments in the glycaemic curve. Consequently, at a glycaemic peak, CRNN yields a predictive curve with even higher slope to compensate the time lag aiming at reducing the gap between the
prediction and real measurements. This feature helps CRNN to decrease the RMSE and MARD as well as maximising the effective prediction horizon.

\subsection{Clinical Data}

\begin{table}[!ht]
	\centering
	\small
	 {
	\caption{RMSE and MARD comparison of different prediction methods for 10 real adult diabetic subjects}\label{table:clinical10RMSE}
	\centering
	\begin{tabular}{|c|c|c|c|c|c|c|}
		\hline
		 \multirow{2}{*}{PH}  & \multirow{2}{*}{Metric} & CRNN & NNPG & SVR & LVX & ARX \\
		\cline{3-7}
		 \multirow{13}{*}{30} &&\multicolumn{5}{c|}{Overall}\\
		\hline
		        & RMSE  & $\textbf{21.07}$ & $23.14$ &$22.00$ & $21.51$ & $21.56$\\
		                       &(mg/dL)&$\pm$2.35&$\pm$2.99&$\pm$2.83&$\pm$2.44&$\pm$2.53\\
		\cline{2-7}
		                       & MARD  & $11.61$ & $13.42$ &$13.54$  & \textbf{$10.93$} & $11.00$  \\
		                       & (\%)  &$\pm$2.18&$\pm$2.35&$\pm$2.88&$\pm$1.87&$\pm$1.81\\
		\cline{2-7}
		                       &       \multicolumn{6}{c|}{Hyperglycaemia}\\
		\cline{2-7}
		                       & MCC & \textbf{$0.79$} & $0.75$ &$0.79$ & $0.79$ & $0.77$  \\
		                       &     & $\pm$0.04&$\pm$0.04&$\pm$0.05&$\pm$0.04&$\pm$0.04\\
		\cline{2-7}
		                       &     \multicolumn{6}{c|}{Hypoglycaemia}\\
		\cline{2-7}
		                       & MCC & $0.51$ & $0.12$\textsuperscript{$\ddagger$} &$0.11$\textsuperscript{$\ddagger$} & $\textbf{0.55}$ & $0.53$  \\
		                       &     & $\pm$0.2&$\pm$0.12&$\pm$0.08&$\pm$0.17&$\pm$0.15\\
		
		\cline{2-7}
		                       &      \multicolumn{6}{c|}{Effective Prediction Horizon}\\
		\cline{2-7}
		                       & Time & $\textbf{19.3}$ & $12.8$\textsuperscript{$\dagger$}&$18.6 $ & $14.5$ & $12.0$\textsuperscript{*}\\
		                       &(min) & $\pm$3.1&$\pm$2.9&$\pm$2.8&$\pm$3.4&$\pm$3.0\\
		\hline
		\multirow{13}{*}{60}&\multicolumn{6}{c|}{Overall}\\
		\cline{2-7}
		& RMSE  & $\textbf{33.27}$ & $36.05$\textsuperscript{$\ddagger$} &$34.35$\textsuperscript{$\dagger$} & $37.46$\textsuperscript{$\ddagger$} & $36.97$\textsuperscript{$\ddagger$}\\
		                       &(mg/dL)&$\pm$4.79&$\pm$4.85&$\pm$4.55&$\pm$5.04&$\pm$4.75\\
		\cline{2-7}
		                       & MARD  & $\textbf{19.01}$ & $21.98$\textsuperscript{$\ddagger$} &$20.65$  & $19.69$\textsuperscript{$\ddagger$} & $19.65$\textsuperscript{$\ddagger$}  \\
		                       & (\%)  &$\pm$4.46&$\pm$4.87&$\pm$3.92&$\pm$3.70&$\pm$3.55\\
		\cline{2-7}
		                       &       \multicolumn{6}{c|}{Hyperglycaemia}\\
		\cline{2-7}
		                       & MCC & $0.72$ & $0.66$\textsuperscript{*} &$0.74$ & $\textbf{0.76}$ & $0.71$  \\
		                       &     & $\pm$0.05&$\pm$0.09&$\pm$0.07&$\pm$0.05&$\pm$0.05\\
		\cline{2-7}
		                       &     \multicolumn{6}{c|}{Hypoglycaemia}\\
		\cline{2-7}
		                       & MCC & $0.40$ & $0.01$\textsuperscript{$\ddagger$} &$0.06$\textsuperscript{$\ddagger$} & $\textbf{0.56}$\textsuperscript{*} & $0.51$  \\
		                       &     & $\pm$0.13&$\pm$ 0.00&$\pm$0.08&$\pm$0.14&$\pm$0.15\\
		
		\cline{2-7}
		                       &      \multicolumn{6}{c|}{Effective Prediction Horizon}\\
		\cline{2-7}
		                       & Time & $\textbf{29.3}$ & $18.3$\textsuperscript{$\ddagger$}&$28.4 $ & $19.9$\textsuperscript{*} & $13.6$\textsuperscript{$\ddagger$}\\
		                       &(min) & $\pm$9.4&$\pm$4.9&$\pm$5.2&$\pm$5.1&$\pm$3.7\\
		\hline	
		\multicolumn{6}{r}{\textsuperscript{*}\scriptsize{$p$-value  $\leq$ 0.05}}  {\textsuperscript{$\dagger$}\scriptsize{$p$-value  $\leq$ 0.01}}  {\textsuperscript{$\ddagger$}\scriptsize{$p$-value $\leq$ 0.005}}\\
	\end{tabular}
		}
	\centering
\end{table}

As mentioned in the previous section, the data obtained in the clinical trial exhibits missing data, and erroneous data. This results in non-physiological discontinuities that would affect the training process. To
mitigate these occurrences, the data is processed with interpolations/extrapolations for gaps in data. The interpolation/extrapolations points are not included in the evaluation of the performance of the methods.

Table \ref{table:clinical10RMSE} shows the RMSE and MARD of the performance of the algorithms for the {10} cases of real data.
 {Contrary to the relative performance of the methods in the in-silico dataset, the evaluation of the methods is mixed}. {The CRNN maintains the best results for RMSE and MARD over a 30 minute prediction horizon baseline
 methods. However, the ARX and LVX models show improved performance in terms of MARD relative to the CRNN. In addition, the LVX shows marginally better performance over CRNN in predicting adverse glycaemic events. The
 time delay in predicting these results shows that CRNN exhibits the best performance with the smallest lag.}

\begin{figure*}[!ht]
     \centering
         \includegraphics[width=4.9in,height=3.2in]{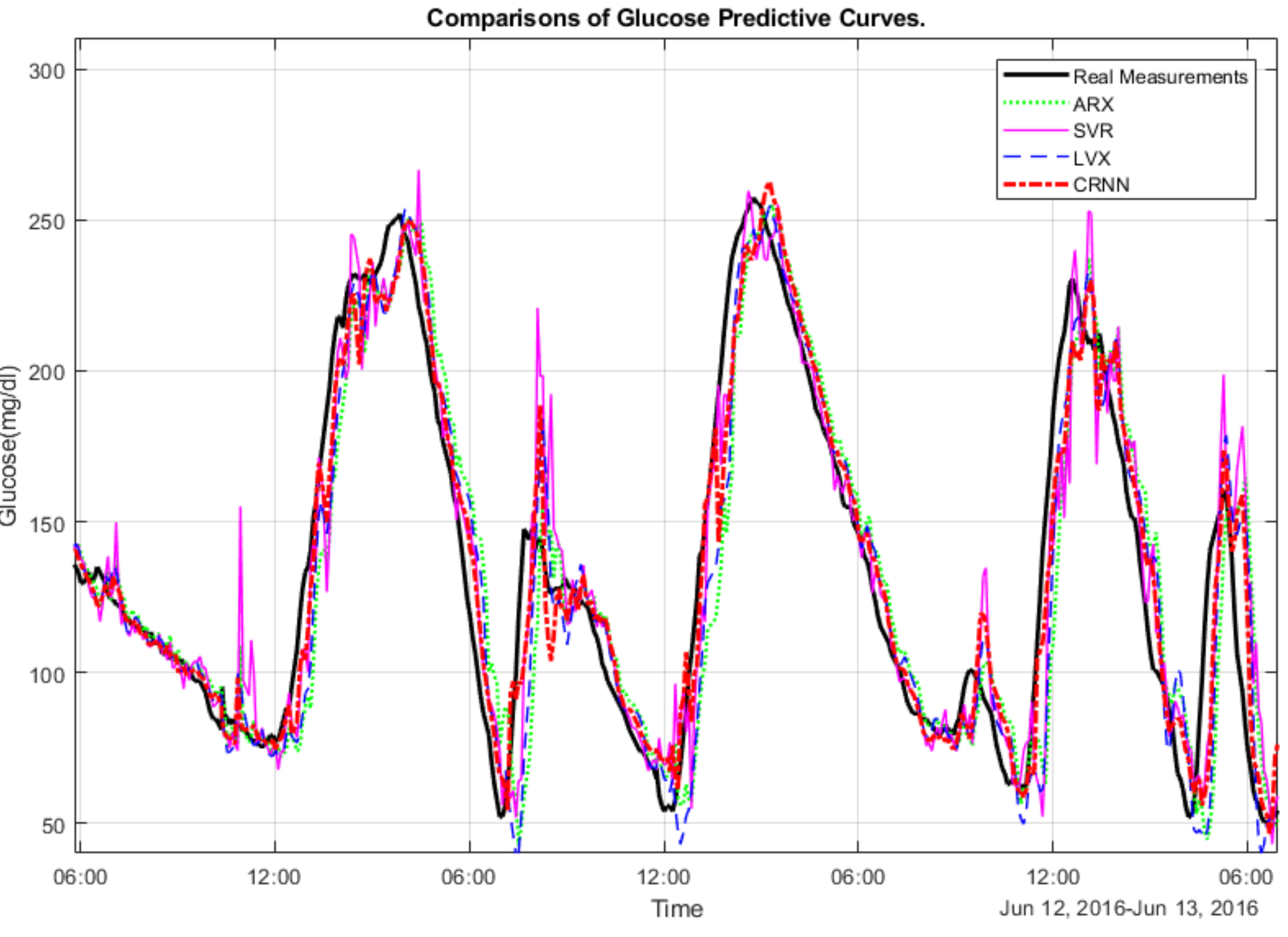}
             \caption{One-day period prediction results for clinical adult 17. The solid black line, dotted green line, solid magenta line, dashed blue line, dash-dotted red line  {indicate the real glucose measurements,
             the prediction results of ARX,  SVR, LVX, and the CRNN algorithm, respectively.} }
             \label{fig_D}
\end{figure*}

 {Over a long-term prediction horizon, the CRNN provides the best performance for prediction of glucose level and with the least lag of the evaluated methods. SVR is able to perform close to the CRNN in terms of effective
 prediction horizon. However, the better prediction of hyperglycaemic events is contrasted with very poor prediction of hypoglycaemia.  The results for hypoglycaemia prediction, equivalent to random guessing, suggest that
 60 minutes represents the limit of meaningful hypoglycaemia prediction for SVR and NNPG given these inputs. Further improvement may require the inclusion of engineered features. Although LVX exhibits superior performance
 in predicting adverse glycaemic events, it should be noted that the user would have considerably less time (-9 mins) to take action.} As seen in Fig. \ref{fig_D}, the CRNN and LVX both achieve good predictive curves
 compared to the ground truth measurements. Specifically, at the inflection periods during peaks and troughs, the LVX tends to have higher and lower predictions, respectively. The CRNN follows the trend at both local and
 global peak points closely, which increases its overall accuracy.

  {
To understand the effect of each network component, we generate networks with different components and evaluate their performances.}
The results are shown in Table \ref{table:SensitivityAnalysis}. It shows that full CRNN achieves the best performance, and both CNN and LSTM component contribute to the final result.  In addition, we investigate the
influence of different lengths of training set.  {The results are in Table \ref{table:TrainingData}.} Using 1 month training data, the RMSE of CRNN can achieve $22.28 \pm 2.67$ (30) and $35.56\pm 4.55$ (60). This can be
slightly improved if longer training data are exploited. It shows that collecting more training data can increase the predictive accuracy.

\begin{table}[ht]
\small
 {
\caption{An ablation study showing the effect of each stage on CRNN performance }\label{table:SensitivityAnalysis}
\centering
\begin{tabular}{|l|c|c|}
\hline
\multirow{2}{4em}{Model} & \multicolumn{2}{|c|}{RMSE} \\
\cline{2-3}
& 30 min  & 60 min\\
\hline
CRNN  & $21.07 \pm 2.35$ & $33.27\pm 4.79$\\
\hline
CRNN w/o CNN  &$22.16 \pm 4.39$ & $36.28\pm 7.14$\\
\hline
CRNN w/o LSTM  & $21.50 \pm 2.62$ & $36.01\pm 6.41$\\
\hline
\end{tabular}
}
\centering
\end{table}

\begin{table}[ht]
\small
 {
\caption{A table showing the performance with different periods of training data }\label{table:TrainingData}
\centering
\begin{tabular}{|l|c|c|}
\hline
\multirow{2}{4em}{Training Data} & \multicolumn{2}{|c|}{RMSE} \\
\cline{2-3}
& 30 min  & 60 min \\
\hline
3 months  & $21.07 \pm 2.35$ & $33.27\pm 4.79$\\
\hline
2 months  & $22.07 \pm 2.84$  & $35.12\pm 4.69$\\
\hline
1 month  & $22.28 \pm 2.67$ & $35.56\pm 4.55$\\
\hline
\end{tabular}
}
\centering
\end{table}

\section{Discussion}
\subsection{Performance in Simulated and Clinical Data}

 In this paper, both \emph{in silico} and clinical dataset are verified. The \emph{in silico} dataset is for test. The clinical dataset, as real data collected from clinical trials, are more practical and significant if
 people want to compare the performances of different methods.
In the previous section we noted a discrepancy in the performance of the proposed algorithm and baseline algorithms in simulated cases and the real patient cases. Previous tests have also indicated that the performance in
real subjects is much less satisfactory comparing to virtual subjects. In our opinion, the drop in performance can be primarily attributed to the increased complexity of real data generated from a patient relative to the
simulated data generated from a physiological model. In addition the gaps in data and method of interpolation/extrapolation may contribute to the further reduction in performance. Relative to the baseline algorithms, the
CRNN is better at capturing the features since deep learning affords a better capacity at learning optimal representations of features. This would explain the relatively lower variance in metrics for the performance of
the CRNN in different cases relative to baseline models.

\subsection{Results Comparisons}
We achieved a mean RMSE $=9.38$mg/dL \emph{in silico} using the proposed method, and it is the best amongst other algorithms, including SVR, LVX and 3rd order ARX. In addition, we want to compare our algorithm with other
approaches in the literature.  {Using dataset generated from simulators,} our algorithm is better than the results of RMSE $= 18.78$mg/dL \cite{Sparacino-GluCon2007} and RMSE = $13.65$mg/dL in
\cite{Mougiakakou-NNbasedGlucose}.
For several other works, it is difficult to evaluate the RMSE through direct comparison  {due to the unavailability of codes, parameters of the models, and the benchmark datasets.} However, we may compare the results with
widely used methods as benchmarks, such as SVR or NNPG.
 For instance, for PH = 30 min as shown in Table 3 \cite{Plis-AMachine2014},  the algorithm is $0.1$ mg/dL better than the result of SVR in terms of RMSE on the real dataset; our algorithm is $0.9$ mg/dL better than the
 SVR in terms of RMSE on the real dataset. In \cite{Zecchin-NNIncor2012}, for PH = 30 min their RMSEs are $1.3$ better than NNPG for the simulated data and $0.2$ mg/dL better than NNPG for the real datasets. Our RMSEs are
 $3.5$ mg/dL better than NNPG for the simulated data and $2.1$ mg/dL better for the real datasets. As far as we know, the proposed algorithm achieves a performance state-of-the-art accuracy with regard to RMSE.
To build a fair comparison, we provide all benchmark models the same input, including CGM data, meal and insulin. For the conventional NNPG, it only uses CGM measurements. Thus in the comparison we incorporates meal and
insulin in the input as well to generate an enhanced NNPG.

\subsection{ {Application on resource-constrained mobile platforms}}
\begin{figure}[!ht]
     \centering
            \includegraphics[width=2.7in,height=2in]{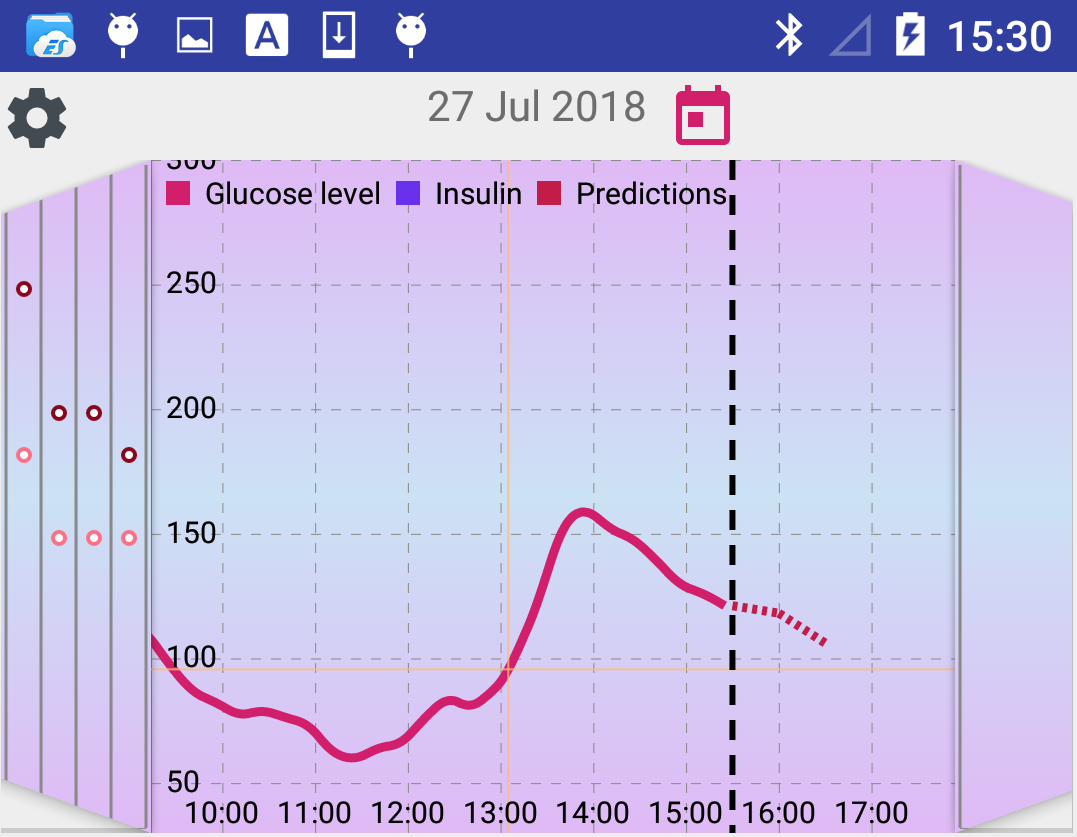}
             \caption{An illustration of the glucose level shown in an app interface on an Android system, where the red curve is the historic blood glucose, black dash line is the current time, and the red dot curve is
             the prediction provided by the model.}\label{fig_E}
\end{figure}

CRNN is a personalised algorithm for different diabetic subjects. Firstly, it is data driven and personalised. Secondly, the model can be continuously updated as more data is available. In details, the model is saved as a
trained neural network. We use the sequential model with Tensorflow backend to train the neural network, and the result can be saved as a small file. {This file can be compiled as a ``.tflite" or a ``.pb" file for the app
on mobiles, by using a Tensorflow Lite converter. The model file can be updated continuously at the cloud.} The app may demonstrate the predictive glycaemic curve on screen. A demonstration on the Android system is shown
in Fig. \ref{fig_E} In addition, we also found that the execution time of the model is $6$ms on a Android phone (LG Nexus5 with Processor: 2.26GHz quad-core, RAM:2GB) and $780$ms on a laptop (MacPro with Processor: 3.1GHz
Intel Core i5, RAM:8GB). The reasons might be in the quantisation of weights and biases (e.g. 8 bit integer vs. 32 bit floating point), thus leading to a faster computation.

\subsection{Limitations}
Though CRNN has achieved very good accuracy in prediction, some challenges remain. Notably the performance in predicting hypoglycaemia degrades much faster than the performance in predicting hyperglycaemia as the
prediction horizon increases. This can relate to the intervention with unaccounted fast acting  {($<15$ mins)} carbohydrates to prevent onset of hypoglycaemia or aerobic exercise which can accelerate onset of
hypoglycaemia. Over farther prediction horizons, future events may need to be accounted for to improve the performance.
In addition,  {meal/bolus rules supported by physiological models might be added to the model.}
Based on the CRNN approach proposed in this paper, it is possible to develop a hybrid method, which may have the advantages of both conventional and  {deep learning} algorithms.

\section{Conclusion}
In this paper a convolutional recurrent neural network was proposed as an effective method for BG prediction. The architecture includes a multi-layer CNN followed by a modified RNN, where the CNN could capture the
features or patterns of the multi-dimensional time series. The modified RNN is capable of analyzing the previous sequential data and providing the predictive BG. The method trains models for each diabetic subject using
their own data.
After obtaining the trained neural network, it could be applied locally or on portable devices. The proposed CRNN method showed superior performance in forecasting BG levels (RMSE and MARD) in the \emph{in silico} and
clinical experiments.
lag.

\section{Acknowledgements}

The authors thank EPSRC, ARISES project for support and providing the clinical data. We would also like to thank Prof. R. Spence, T. Zhu and C. Demasson's for the contribution to the mobile app interface design. 

\section{Appendix}
\begin{table}[!h]
\centering
 {
	\small
	\caption{A Table detailing the size and dimensions of layers in CRNN}\label{table:Adult10RMSE}
	\centering
	\begin{tabular}{|c|c|c|}
		\hline
		Layer Description & Output DImensions & No. of \\
		(layer) & &Parameters\\
		\hline
		\multicolumn{3}{|c|}{Convolutional Layers (Batch$\times$Steps$\times$Channels)}\\
		\hline
		(1) 1$\times$4 conv &$ 128 (1) \times 24 \times 8 $ & $104$ \\
		\hline
		 max\_pooling, size 2 &$ 128 (1) \times 12 \times 8$ &$-$ \\
		\hline
		(2) 1$\times$4 conv &$ 128 (1) \times 12 \times 16$ &$528$ \\
		\hline
		 max\_pooling, size 2 &$ 128 (1) \times 6 \times 16$ &$-$ \\
		\hline
		(3) 1$\times$4 conv &$ 128 (1) \times 6 \times 32$ &$2080$ \\
		\hline
		 max\_pooling &$ 128 (1) \times 3 \times 32 $ &$-$ \\
		\hline
		\multicolumn{3}{|c|}{Recurrent Layer (Batch$\times$Cells)}\\
		\hline
		(4) lstm &$ 128 (1) \times 64 $ &$24832$ \\
		\hline
		\multicolumn{3}{|c|}{Dense Layers (Batch$\times$Units)}\\
		\hline
		(5) dense &$ 128 (1) \times 256$ &$16640$ \\
		\hline
		(6) dense &$ 128 (1) \times 32$ & $8224$\\
		\hline
		(7) dense &$128 (1) \times 1$ &$33$ \\
		\hline
	\end{tabular}
	\centering
	}
\end{table}

\bibliographystyle{IEEEtran}

\end{document}